\documentclass[10pt,twocolumn,letterpaper]{article}

\usepackage{iccv}
\usepackage{times}
\usepackage{epsfig}
\usepackage{graphicx}
\usepackage{amsmath}
\usepackage{amssymb}
\usepackage{multirow}
\usepackage{graphicx}
\usepackage{subfigure}
\usepackage{color}

\usepackage[pagebackref=true,breaklinks=true,letterpaper=true,colorlinks,bookmarks=false]{hyperref}

\newcommand{\bx} {{\bf x }}

\newcommand{\bh} {{\bf h }}

\newcommand{\by} {{\bf y }}

\newcommand{\bI} {{\bf I }}
\iccvfinalcopy 

\def\iccvPaperID{677} 

\ificcvfinal\fi

\begin{document}

\title{From Facial Parts Responses to Face Detection: A Deep Learning Approach}

\author{Shuo Yang$^{1,2}$\\\and Ping Luo$^{2,1}$\\\and Chen Change Loy$^{1,2}$\\\and Xiaoou Tang$^{1,2}$\\\and
\small{$^1$Department of Information Engineering, The Chinese University of Hong Kong}\\
\small{$^2$Shenzhen Key Lab of Comp. Vis. \& Pat. Rec., Shenzhen Institutes of Advanced Technology, CAS, China}\\
{\tt\small \{ys014, pluo, ccloy, xtang\}@ie.cuhk,edu.hk}
}

\maketitle

\begin{abstract}

In this paper, we propose a novel deep convolutional network (DCN) that achieves outstanding performance on FDDB, PASCAL Face, and AFW. Specifically, our method achieves a high recall rate of 90.99\% on the challenging FDDB benchmark, outperforming the state-of-the-art method~\cite{HeadHunter} by a large margin of 2.91\%.
Importantly, we consider finding faces from a new perspective through scoring facial parts responses by their spatial structure and arrangement. The scoring mechanism is carefully formulated considering challenging cases where faces are only partially visible. This consideration allows our network to detect faces under severe occlusion and unconstrained pose variation, which are the main difficulty and bottleneck of most existing face detection approaches.
We show that despite the use of DCN, our network can achieve practical runtime speed.

\end{abstract}

\section{Introduction}
\label{sec:introduction}

\begin{figure}[h]
\begin{center}
\includegraphics[width=\linewidth]{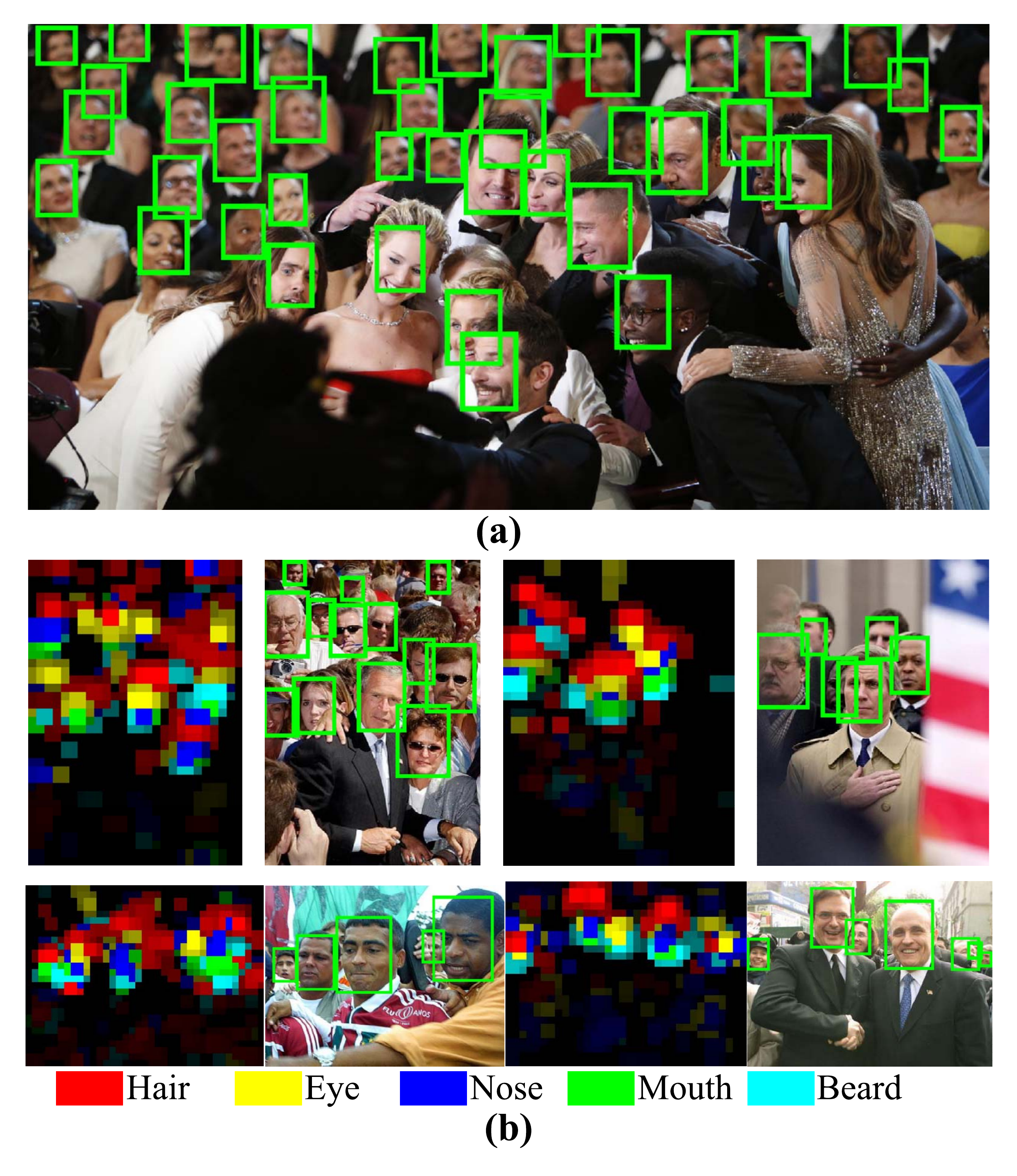}
\vskip -0.3cm
\caption{(a) We propose a deep convolutional network for face detection, which achieves high recall of faces even under severe occlusions and head pose variations. The key to the success of our approach is the new mechanism for scoring face likeliness based on deep network responses on local facial parts. (b) The part-level response maps (we call it `partness' map) generated by our deep network given a full image without prior face detection. All these occluded faces are difficult to handle by conventional approach.}
\label{fig:intro}
\end{center}
\vskip -0.5cm
\end{figure}

Neural network based methods were once widely applied for localizing faces~\cite{vaillant1994original,rowley1998neural,garcia2004convolutional,osadchy2007synergistic}, but they were soon replaced by various non-neural network-based face detectors, which are based on cascade structure~\cite{JointCascade,huang2007high,SURF,viola2004robust} and deformable part models (DPM)~\cite{HeadHunter,yan2014face,zhu2012face} detectors.
%
%
Deep convolutional networks (DCN) have recently achieved remarkable performance in many computer vision tasks, such as object detection, object classification, and face recognition.
Given the recent advances of deep learning and graphical processing units (GPUs), it is worthwhile to revisit the face detection problem from the neural network perspective.

In this study, we wish to design a deep convolutional network for face detection, with the aim of not only exploiting the representation learning capacity of DCN, but also formulating a novel way for handling the severe occlusion issue, which has been a bottleneck in face detection.
To this end, we design a new deep convolutional network with the following appealing properties:
(1) It is robust to severe occlusion. As depicted in Fig.~\ref{fig:intro}, our method can detect faces even more than half of the face region is occluded; (2) it is capable of detecting faces with large pose variation, \eg~profile view without training separate models under different viewpoints; (3) it accepts full image of arbitrary size and the faces of different scales can appear anywhere in the image.

All the aforementioned properties, which are challenging to achieve with conventional approaches, are made possible with the following considerations:

\noindent
(1) \textit{Generating face parts responses from attribute-aware deep networks}:
We believe the reasoning of unique structure of local facial parts (\eg eyes, nose, mouths) is the key to address face detection in unconstrained environment. To this end, we design a set of attribute-aware deep networks, which are pre-trained with generic objects and then fine-tuned with specific part-level binary attributes (\eg~mouth attributes including big lips, opened mouth, smiling, wearing lipstick). We show that these networks could generate response maps in deep layers that strongly indicate the locations of the parts.
The examples depicted in Fig.~\ref{fig:intro}(b) show the responses maps (known as `partness map' in our paper) of five different face parts.

\noindent
(2) \textit{Computing faceness score from responses configurations}:
Given the parts responses, we formulate an effective method to reason the degree of face likeliness through analysing their spatial arrangement. For instance, the hair should appear above the eyes, and the mouth should only appear below the nose. Any inconsistency would be penalized. Faceness scores will be derived and used to re-rank candidate windows of any generic object proposal generator to obtain a set of face proposals.  Our experiment shows that our face proposal enjoys a high recall with just modest number of proposals (over 90\% of face recall with around $150$ proposals, $\approx$0.5\% of full sliding windows, and $\approx$10\% of generic object proposals).

\noindent
(3) \textit{Refining the face hypotheses} --
Both the aforementioned components offer us the chance to find a face even under severe occlusion and pose variations.
The output of these components is a small set of high-quality face bounding box proposals that cover most faces in an image.
Given the face proposals, we design a multitask deep convolutional network in the second stage to refine the hypotheses further, by simultaneously recognizing the true faces and estimating more precise face locations.

Our main contribution in this study is the novel use of DCN for discovering facial parts responses from arbitrary uncropped face images. Interestingly,  
in our method, part detectors emerge within CNN trained to classify attributes from uncropped face images, without any part supervision.
This is new in the literature. We leverage this new capability to further propose a face detector that is robust to severe occlusion. Our network achieves the state-of-the-art performance on challenging face detection benchmarks including FDDB, PASCAL Faces, and AFW. We show that practical runtime speed can be achieved albeit the use of DCN.

\section{Related Work}
\label{sec:related_work}

There is a long history of using neural network for the task of face detection~\cite{vaillant1994original,rowley1998neural,garcia2004convolutional,osadchy2007synergistic}.
An early face detection survey~\cite{yang2002detecting} provides an extensive coverage on relevant methods. Here we highlight a few notable studies.
Rowley~\etal~\cite{rowley1998neural} exploit a set of neural network-based filters to detect presence of faces in multiple scales, and merge the detections from individual filters.
Osadchy~\etal~\cite{osadchy2007synergistic} demonstrate that a joint learning of face detection and pose estimation significantly improves the performance of face detection.
The seminal work of Vaillant~\etal~\cite{vaillant1994original} adopt a two-stage coarse-to-fine detection. Specifically, the first stage approximately locates the face region, whilst the second stage provides a more precise localization.
Our approach is inspired by these studies, but we introduce innovations on many aspects.
In particular, we employ contemporary deep learning strategies, \eg~pre-training, to train deeper networks for more robust feature representation learning. Importantly, our first stage network is conceptually different from that of~\cite{vaillant1994original}, and many recent deep learning detection frameworks -- we train attribute-aware deep convolutional networks to achieve precise localization of facial parts, and exploit their spatial structure for inferring face likeliness. This concept is new and it allows our model to detect faces under severe occlusion and pose variations.
While great efforts have been devoted for addressing face detection under occlusion~\cite{lin2005robust,lin2004fast}, these methods are all confined to frontal faces. In contrast, our model can discover faces under variations of both pose and occlusion.

In the last decades, cascade based~\cite{JointCascade,huang2007high,SURF,viola2004robust} and deformable part models (DPM) detectors dominate the face detection approaches. Viola and Jones~\cite{viola2004robust} introduced fast Haar-like features computation via integral image and boosted cascade classifier. Various studies thereafter follow a similar pipeline. Amongst the variants, SURF cascade~\cite{SURF} was one of the top performers.
Later Chen~\etal~\cite{JointCascade} demonstrate state-of-the-art face detection performance by learning face detection and face alignment jointly in the same cascade framework.
Deformable part models define face as a collection of parts. Latent Support Vector Machine is typically used to find the parts and their relationships. DPM is shown more robust to occlusion than the cascade based methods.
A recent study~\cite{HeadHunter} demonstrates state-of-the-art performance with just a vanilla DPM, achieving better results than more sophisticated DPM variants~\cite{yan2014face,zhu2012face}.

\begin{figure*}[t]
\begin{center}
{\includegraphics[width=\linewidth]{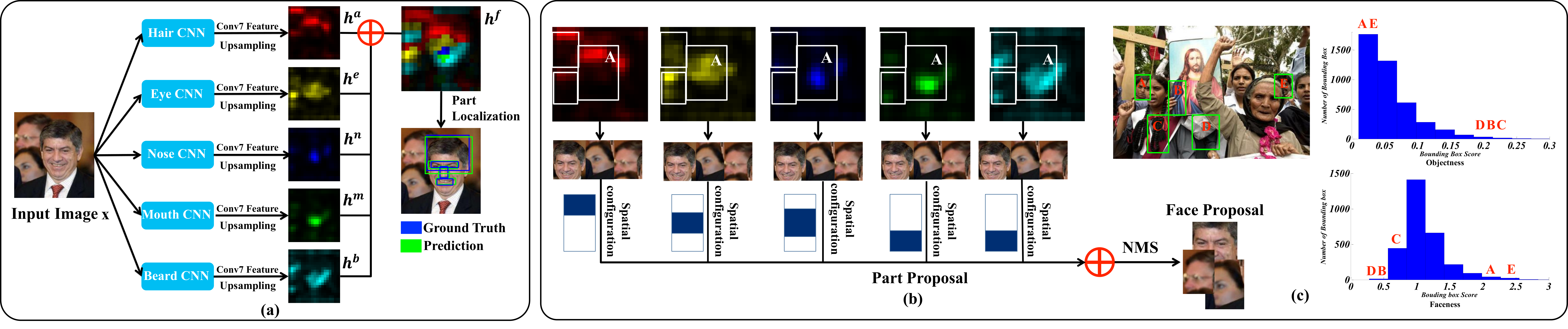}}
{\caption{\small{(a) The pipeline of generating part response maps and part localization. Different CNNs are trained to handle different facial parts, but they can share deep layers for computational efficiency. (b) The pipeline for generating face proposals. (c) Bounding box reranking by face measure \bf{(Best viewed in color)}.}}\label{fig:pipeline_1}}
\vspace{-0.45cm}
\end{center}
\end{figure*}

A recent study~\cite{farfade2015multi} shows that face detection can be further improved by using deep learning, leveraging the high capacity of deep convolutional networks. In this study, we push the performance limit further.
Specifically, the network proposed by~\cite{farfade2015multi} does not have explicit mechanism to handle occlusion, the face detector therefore fails to detect faces with heavy occlusions, as acknowledged by the authors. In contrast, our two-stage architecture has its first stage designated to handle partial occlusions. In addition,
our network gains improved efficiency by adopting the more recent fully convolutional architecture, in contrast to the previous work that relies on the conventional sliding window approach to obtain the final face detector.

The first stage of our model is partially inspired by the generic object proposal approaches~\cite{arbelaez2014multiscale,uijlings2013selective,zitnick2014edge}.
Generic object proposal generators are now an indispensable component of standard object detection algorithms through providing high-quality and category-independent bounding boxes.
These generic methods, however, are devoted to generic objects therefore not suitable to propose windows specific to face.
In particular, applying a generic proposal generator directly would produce enormous number of candidate windows but only minority of them contain faces.
In addition, a generic method does not consider the unique structure and parts on the face. Hence, there will be no principled mechanism to recall faces when the face is only partially visible.
These shortcomings motivate us to formulate the new faceness measure to achieve high recall on faces, whilst reduce the number of candidate windows to half the original.

\section{Faceness-Net}
\label{sec:methodology}

This section introduces the proposed attribute-aware face proposal and face detection approach, \textit{Faceness-Net}.
In the following, we first briefly overview the entire pipeline and then discuss the details.

Faceness-Net's pipeline consists of three stages, \ie generating partness maps, ranking candidate windows by faceness scores, and refining face proposals for face detection.
In the first stage as shown in Fig.~\ref{fig:pipeline_1}(a), a full image $\bx$ is used as input to five CNNs.  Note that all the five CNNs can share deep layers to save computational time.
Each CNN outputs a partness map, which is obtained by weighted averaging over all the label maps at its top convolutional layer.
Each of these partness maps indicates the location of a specific facial component presented in the image, \eg hair, eyes, nose, mouth, and beard,
denoted by $\bh^a$, $\bh^e$, $\bh^n$, $\bh^m$, and $\bh^b$, respectively. We combine all these partness maps into a face label map $\bh^f$, which clearly designates faces' locations.

In the second stage, given a set of candidate windows that are generated by existing object proposal methods such as \cite{arbelaez2014multiscale,uijlings2013selective,zitnick2014edge},
we rank these windows according to their faceness scores, which are extracted from the partness maps with respect to different facial parts configurations, as illustrated at the bottom of Fig.~\ref{fig:pipeline_1}(b).
For example, as visualized in Fig.~\ref{fig:pipeline_1}(b), a candidate window `A' covers a local region of $\bh^a$ (\ie hair) and its faceness score is calculated by dividing the values at its upper part with respect to the values at its lower part, because hair is more likely to present at the top of a face region.
A final faceness score of `A' is obtained by averaging over the scores of these parts.
In this case, large number of false positive windows can be pruned.
Notably, the proposed approach is capable of coping with severe face occlusions, as shown in Fig.~\ref{fig:pipeline_1}(c), where face windows `A' and `E' can be retrieved by objectness~\cite{alexe2012measuring} only if large amount of windows are proposed, whilst they rank top $50$ by using our method.

In the last stage, the proposed candidate windows are refined by training a multitask CNN, where face classification and bounding box regression are jointly optimized.

\subsection{Partness Maps Extraction}
\label{sec:partness}

\begin{figure}[t]
\begin{center}
\vskip -0.1cm
\includegraphics[width=\linewidth]{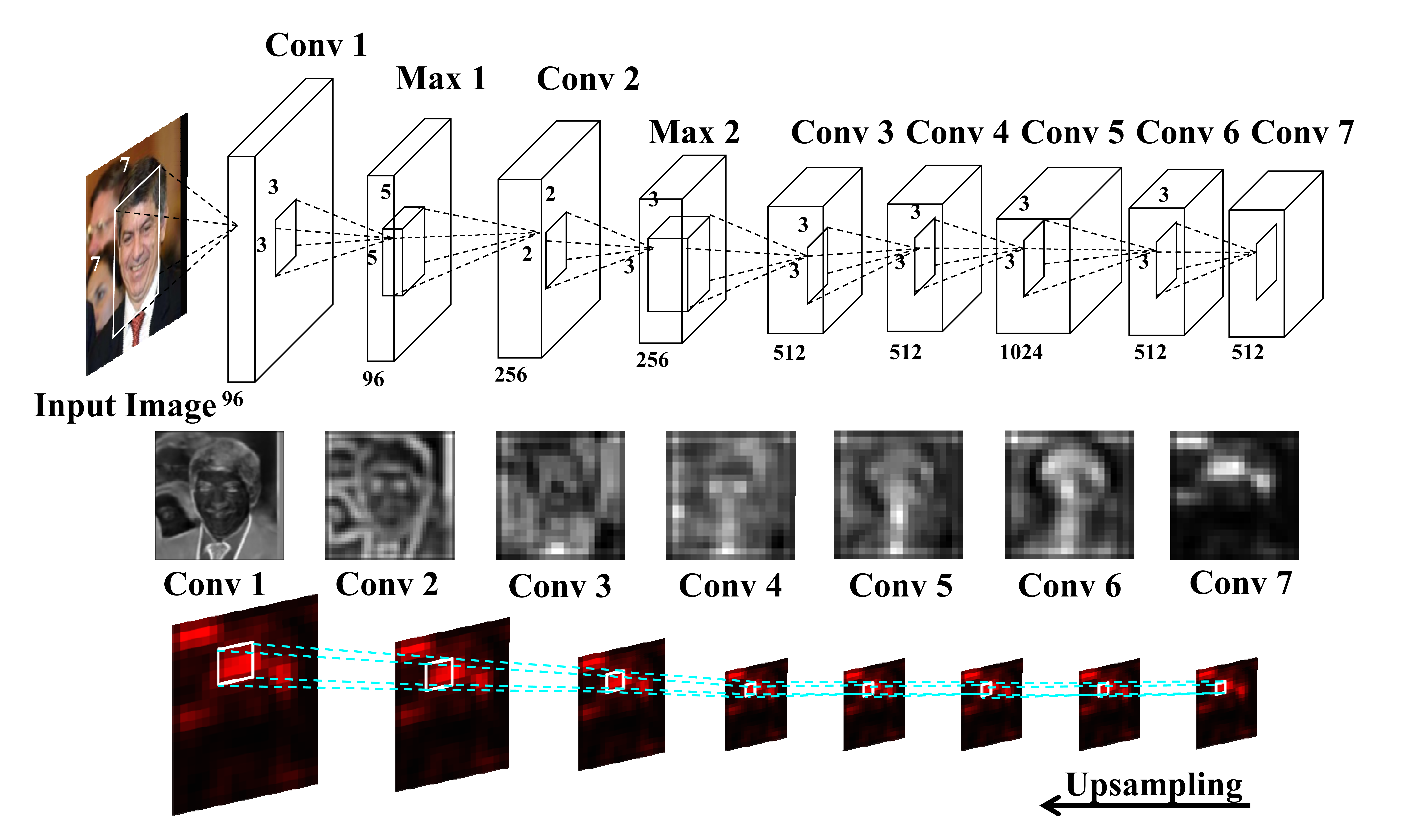}
\caption{A general architecture of an attribute-aware deep network. Other architecture is possible.}
\label{fig:network}
\end{center}
\vspace{-0.45cm}
\end{figure}

\noindent\textbf{Network structure}. Fig.~\ref{fig:network} depicts the structure and hyper-parameters of the CNN in Fig.~\ref{fig:pipeline_1}(a), which stacks seven convolutional layers (conv1 to conv7) and two max-pooling layers (max1 and max2).
This convolutional structure is inspired by the AlexNet~\cite{imagenet} in image classification. Many recent studies~\cite{simonyan2013deep, clarifai14} showed that stacking many convolutions as AlexNet did can roughly capture object locations.

\begin{figure}[h]
\begin{center}
\includegraphics[width=\linewidth]{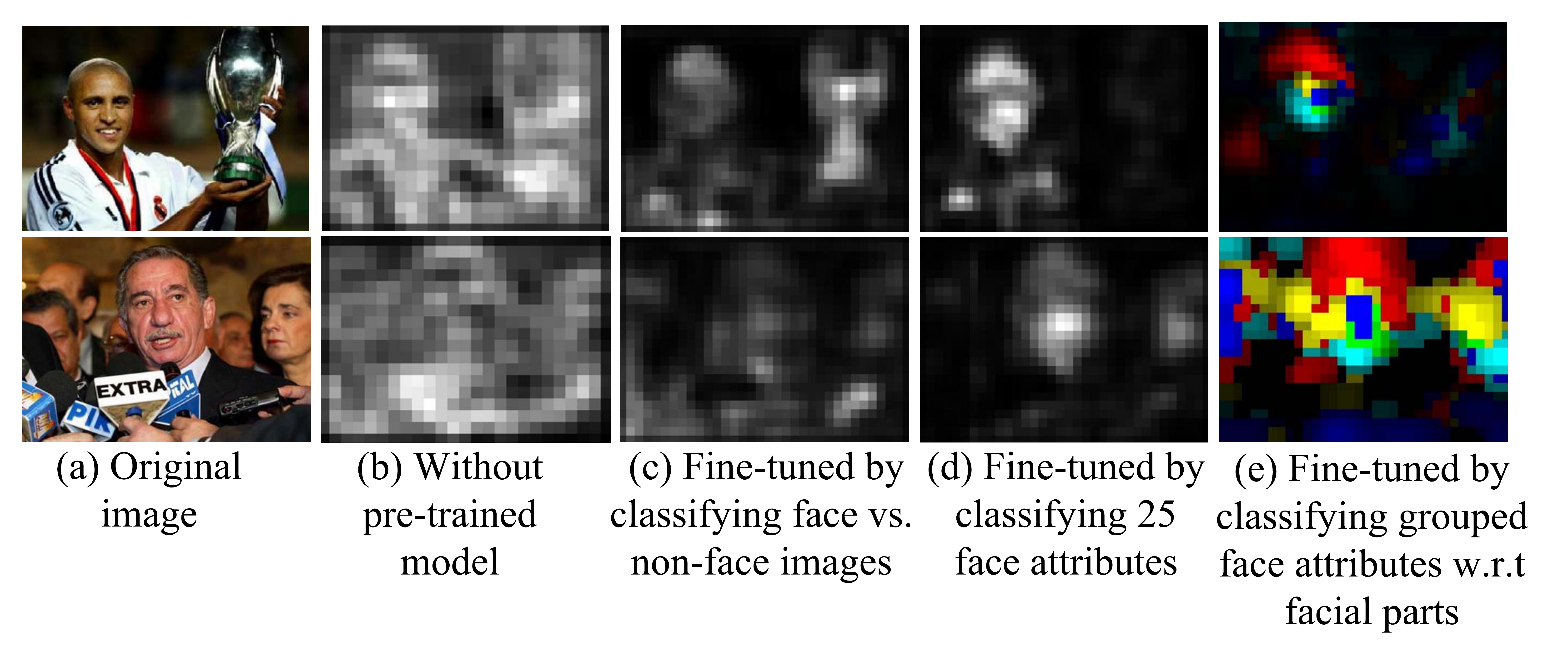}
\vskip -0.1cm
\caption{The responses or partness maps obtained by using different types of supervisions.}
\label{fig:partness_map_learning}
\end{center}
\vspace{-0.35cm}
\end{figure}

\begin{table}
\begin{center}
\caption{Facial attributes grouping.}
\vspace{0.05cm}
\label{tab:attribute_category}
\footnotesize
\addtolength{\tabcolsep}{-1pt}
\begin{tabular}{|c||c|}
\hline
\textbf{Facial Part} & \textbf{Facial Attributes}
\\
\hline
\hline
Hair & Black hair, Blond hair, Brown hair, Gray hair, Bald, \\& Wavy hair, Straight hair, Receding hairline, Bangs\\
\hline
Eye & Bushy eyebrows, Arched eyebrows, Narrow eyes,\\ & Bags under eyes, Eyeglasses \\
\hline
Nose & Big nose, Pointy nose \\
\hline
Mouth &	Big lips, Mouth slightly open, Smiling,\\ & Wearing lipstick\\
\hline
Beard &	No beard, Goatee, 5 o'clock shadow, \\ & Mustache, Sideburns\\
\hline
\end{tabular}
\end{center}
\vskip -0.45cm
\vspace{-0.4cm}
\end{table}

\noindent \textbf{Learning partness maps}.
As shown in Fig.~\ref{fig:partness_map_learning}(b), a deep network trained on generic objects, \eg~AlexNet~\cite{imagenet}, is not capable of providing us with precise faces' locations, let alone partness map.
The partness maps can be learned in multiple ways.
The most straight-forward manner is to use the image and its pixelwise segmentation label map as input and target, respectively.
This setting is widely employed in image labeling~\cite{farabet2013learning,aerial_hinton}.
However, it requires label maps with pixelwise annotations, which are expensive to collect.
Another setting is image-level classification (\ie faces and non-faces), as shown in Fig.~\ref{fig:partness_map_learning}(c). It works well where the training images are well-aligned, such as face recognition~\cite{sun2014deepb}.
Nevertheless, it suffers from complex background clutter because the supervisory information is not sufficient to account for face variations.
Its learned feature maps contain too much noises, which overwhelm the actual faces' locations.
Attribute learning in Fig.~\ref{fig:partness_map_learning}(d) extends the binary classification in (c) to the extreme by using a combination of attributes to capture face variations.
For instance, an `Asian' face can be distinguished from a `European' face.
However, our experiments demonstrate that the setting is not robust to occlusion.
Hence, as shown in Fig.~\ref{fig:partness_map_learning}(e), this work extends (d) by partitioning attributes into groups based on facial components.
For instance, `black hair', `blond hair', `bald', and `bangs' are grouped together, as all of them are related to hair.
The grouped attributes are summarized in Table~\ref{tab:attribute_category}.
In this case, different face parts can be modeled by different CNNs (with option to share some deep layers). If one part is occluded, the face region can still be localized by CNNs of the other parts.

We take the Hair-CNN in Fig.~\ref{fig:pipeline_1}(a) as an example to illustrate the learning procedure.
Let $\{\bx_i,\by_i\}_{i=1}^N$ be a set of full face images and the attribute labels of hair, where $\forall\bx_i\in\mathbb{R}^{256\times256}$ and $\forall\by_i\in\mathbb{R}^{1\times 9}$, implying that each full image is rescaled to $256\times256$ and there is nine attributes related to hair as listed in Table~\ref{tab:attribute_category}.
Learning is formulated as a multi-variate classification problem by minimizing the cross-entropy loss,
$L=\sum_{i=1}^{N}\by_i\log p(\mathbf{y}_i=1|\bx_i) + ({\mathbf{1}} - \by_i)\log\big({\mathbf{1}} - p(\by_i=1|\bx_i)\big)$,
where $p(\by_i|\bx_i)$ is modeled as a sigmoid function,
indicating the probability of the presence of the attributes.
This loss function can be optimized by the stochastic gradient descent with back-propagation.

However, the partness map generated by the Hair-CNN trained as above contains erroneous responses at the background, revealing that this training scheme is not sufficient to account for background clutter. To obtain a cleaner partness map, we employ the merit from object categorization, where CNN is pre-trained with massive general object categories in ImageNet~\cite{russakovsky2014imagenet} as in \cite{imagenet}. It can be viewed as a supervised pre-training for the Hair-CNN.

\subsection{Ranking Windows by Faceness Measure}
\label{sec:faceness_measure}

\begin{figure}[t]
\begin{center}
{\includegraphics[width=7cm]{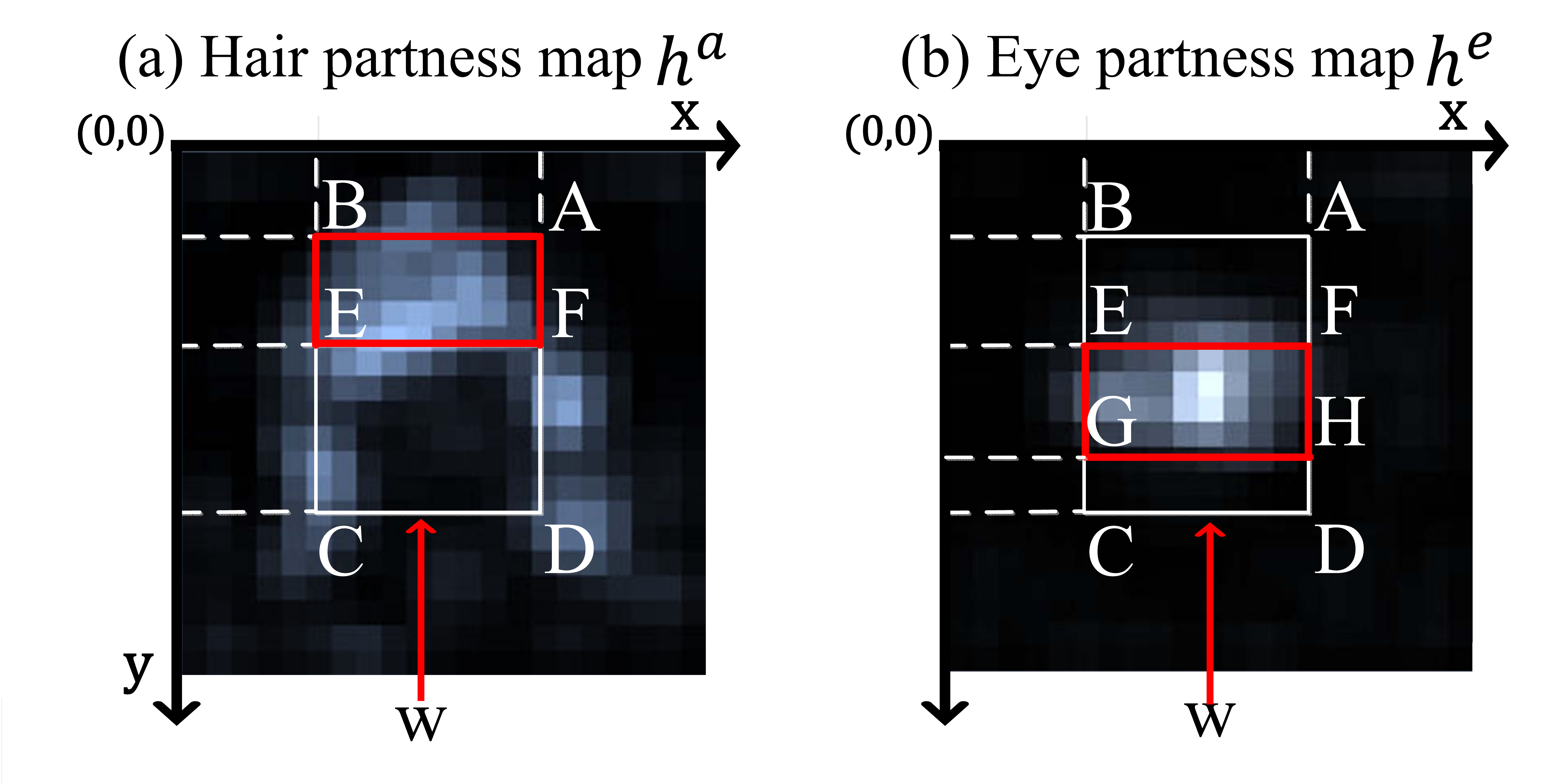}}
{\caption{\small{ Examples of spatial configurations (\textbf{Best viewed in color}).}}\label{fig:integral}}
\vspace{-0.45cm}
\end{center}
\end{figure}

Our approach is loosely coupled with existing generic object proposal generators~\cite{arbelaez2014multiscale,uijlings2013selective,zitnick2014edge} - it accepts candidate windows from the latter but generates its own faceness measure to return a ranked set of top-scoring face proposals.
Fig.~\ref{fig:integral} takes hair and eyes to illustrate the procedure of deriving the faceness measure from a partness map.
Let $\Delta_w$ be the faceness score of a window $w$.
For example, as shown in Fig.~\ref{fig:integral}(a), given a partness map of hair, $\bh^a$, $\Delta_w$ is attained by dividing the sum of values in ABEF (red) by the sum of values in FECD.
Similarly, Fig.~\ref{fig:integral}(b) expresses that $\Delta_w$ is obtained by dividing the sum of values in EFGH (red) with respect to ABEF+HGCD of $\bh^e$.

For both of the above examples, larger value of $\Delta_w$ indicates $w$ has higher overlapping ratio with face.
These spatial configurations, such as ABEF in (a) and EFGH in (b), can be learned from data.
We take hair as an example.
We need to learn the positions of points E and F, which can be represented by the $(x,y)$-coordinates of ABCD, \ie the proposed window.
For instance, the position of E in (a) can be represented by $x_e=x_b$ and $y_e=\lambda y_b+(1-\lambda)y_c$, implying that the value of its $y$-axis is a linear combination of $y_b$ and $y_c$.
With this representation, $\Delta_w$ can be efficiently computed by using the integral image (denoted as $\bI$) of the partness map. For instance, $\Delta_w$ in (a) is attained by
\begin{equation}
\frac{\bI(x_f,y_f)+\bI(x_b,y_b)-\bI(x_a,y_a)-\bI\big(x_b,\lambda y_b+(1-\lambda)y_c\big)}{\bI(x_d,y_d)+\bI(x_e,y_e)-\bI\big(x_a,\lambda y_a+(1-\lambda)y_d\big)-\bI(x_c,y_c)},
\end{equation}
where $\bI(x,y)$ signifies the value at the location $(x,y)$.

Given a training set $\{w_i,r_i,\bh_i\}_{i=1}^M$, where $w_i$ and $r_i\in\{0,1\}$ denote the $i$-th window and its label (\ie face/non-face), respectively.
$\bh_i$ is the cropped partness map with respect to the $i$-th window, \eg region ABCD in $\bh^a$. This problem can be simply formulated as maximum a posteriori (MAP)
\begin{equation}\label{eq:map}
\lambda^\ast=\arg\max_{\lambda}\prod_i^M p(r_i|\lambda,w_i,\bh_i)p(\lambda,w_i,\bh_i),
\end{equation}
where $\lambda$ represents a set of parameters when learning the spatial configuration of hair (Fig.~\ref{fig:integral}(a)).
$p(r_i|\lambda,w_i,\bh_i)$ and $p(\lambda,w_i,\bh_i)$ stand for the likelihood and prior, respectively. The likelihood of faceness can be modeled by a sigmoid function, \ie $p(r_i|\lambda,w_i,\bh_i)=\frac{1}{1+\exp(\frac{-\alpha}{\Delta_{w_i}})}$, where $\alpha$ is a coefficient.
This likelihood measures the confidence of partitioning face and non-face, given a certain spatial configuration.
The prior term can be factorized, $p(\lambda,w_i,\bh_i)=p(\lambda)p(w_i)p(\bh_i)$,
where $p(\lambda)$ is a uniform distribution between zero and one, as it indicates the coefficients of linear combination, $p(w_i)$ models the prior of the candidate window, which can be generated by object proposal methods, and $p(\bh_i)$ is the partness map as in Sec.~\ref{sec:partness}. 
Since $\lambda$ typically has low dimension in this work (\eg one dimension of hair), it can be simply obtained by line search. Nevertheless, Eq.(\ref{eq:map}) can be easily extended to model more complex spatial configurations.

\subsection{Face Detection}
\label{sec:face_detection}
The proposed windows achieved by faceness measure have high recall rate.
To improve it further, we refine these windows by joint training face classification and bounding box regression using a CNN similar to the AlexNet~\cite{imagenet}.

In particular, we fine-tune AlexNet using face images from AFLW~\cite{aflw} and person-free images from PASCAL VOC 2007~\cite{pascalvoc}.
For face classification, a proposed window is assigned with a positive label if the IoU between it and the ground truth bounding box is larger than $0.5$; otherwise it is negative.
For bounding box regression, each proposal is trained to predict the positions of its nearest ground truth bounding box.
If the proposed window is a false positive, the CNN outputs a vector of $[-1,-1,-1,-1]$.
We adopt the Euclidean loss and cross-entropy loss for bounding box regression and face classification, respectively.
%

\section{Experimental Settings}
\label{sec:settings}

\noindent \textbf{Training datasets}.
\noindent (i) We employ CelebFaces dataset~\cite{celeface} to train our attribute-aware networks. The dataset contains 87,628 web-based images exclusive from the LFW~\cite{lfw}, FDDB~\cite{fddb}, AFW~\cite{zhu2012face} and PASCAL~\cite{yan2014face} datasets.
We label all images in the CelebFaces dataset with $25$ facial attributes and divide the labeled attributes into five categories based on their respective facial parts as shown in Table~\ref{tab:attribute_category}.
We randomly select $75,000$ images from the CelebFaces dataset for training and the remaining is reserved as validation set.
%
\noindent (ii)  
For face detection training, we choose $13,205$ images from the AFLW dataset~\cite{aflw} to ensure a balanced out-of-plane pose distribution and $5,771$ random person-free images from the PASCAL VOC $2007$ dataset.

\noindent \textbf{Part response testing dataset}.
In Sec.~\ref{sec:evaluate_part_localisation}, we use LFW dataset~\cite{lfw} for evaluating the quality of part response maps for part localization.
We select $2,927$ LFW images following~\cite{crf} since it provides manually labeled hair+beard superpixel labels, on which the minimal and maximal coordinates can be used to generate the ground truth of face parts bounding boxes. Similarly, face parts boxes for eye, nose and mouth are manually labeled guided by the $68$ dense facial landmarks.

\noindent \textbf{Face proposal and detection testing datasets}.
In Sec.~\ref{sec:evaluate_face_proposal} and Sec.~\ref{sec:evaluate_face_detection}, we use the following datasets.
\noindent (i) FDDB~\cite{fddb} dataset contains the annotations for $5,171$ faces in a set of $2,845$ images. For the face proposal evaluation, we follow the standard evaluation protocol in object proposal studies~\cite{zitnick2014edge} and transform the original FDDB ellipses ground truth into bounding boxes by minimal bounding rectangle.
For the face detection evaluation, the original FDDB ellipse ground truth is used.
\noindent (ii) AFW~\cite{zhu2012face} dataset is built using Flickr images. It has $205$ images with $473$ annotated faces with large variations in both face viewpoint and appearance.
\noindent (iii) PASCAL faces~\cite{yan2014face} is a widely used face detection benchmark dataset. It consists of $851$ images and $1,341$ annotated faces.

\noindent \textbf{Evaluation settings}.
Following~\cite{zitnick2014edge}, we employ the Intersection over Union (IoU) as evaluation metric.
We fix the IoU threshold to $0.5$ following the strict PASCAL criterion. In particular, an object is considered being covered/detected by a proposal if IoU is no less than $0.5$.
To evaluate the effectiveness of different object proposal algorithms, we use the detection rate (DR) given the number of proposals per image~\cite{zitnick2014edge}.
For face detection, we use standard precision and recall (PR) to evaluate the effectiveness of face detection algorithms.
\section{Results}
\label{sec:experiments}

\subsection{Evaluating the Quality of Partness Maps}
\label{sec:evaluate_part_localisation}

\noindent \textbf{Robustness to unconstrained training input.}
In the testing stage, the proposed approach does not assume well-cropped faces as input. In the training stage, our approach neither requires well-cropped faces for learning. This is an unique advantage over existing approaches.

\begin{figure}[t]
\begin{center}
\includegraphics[width=\linewidth]{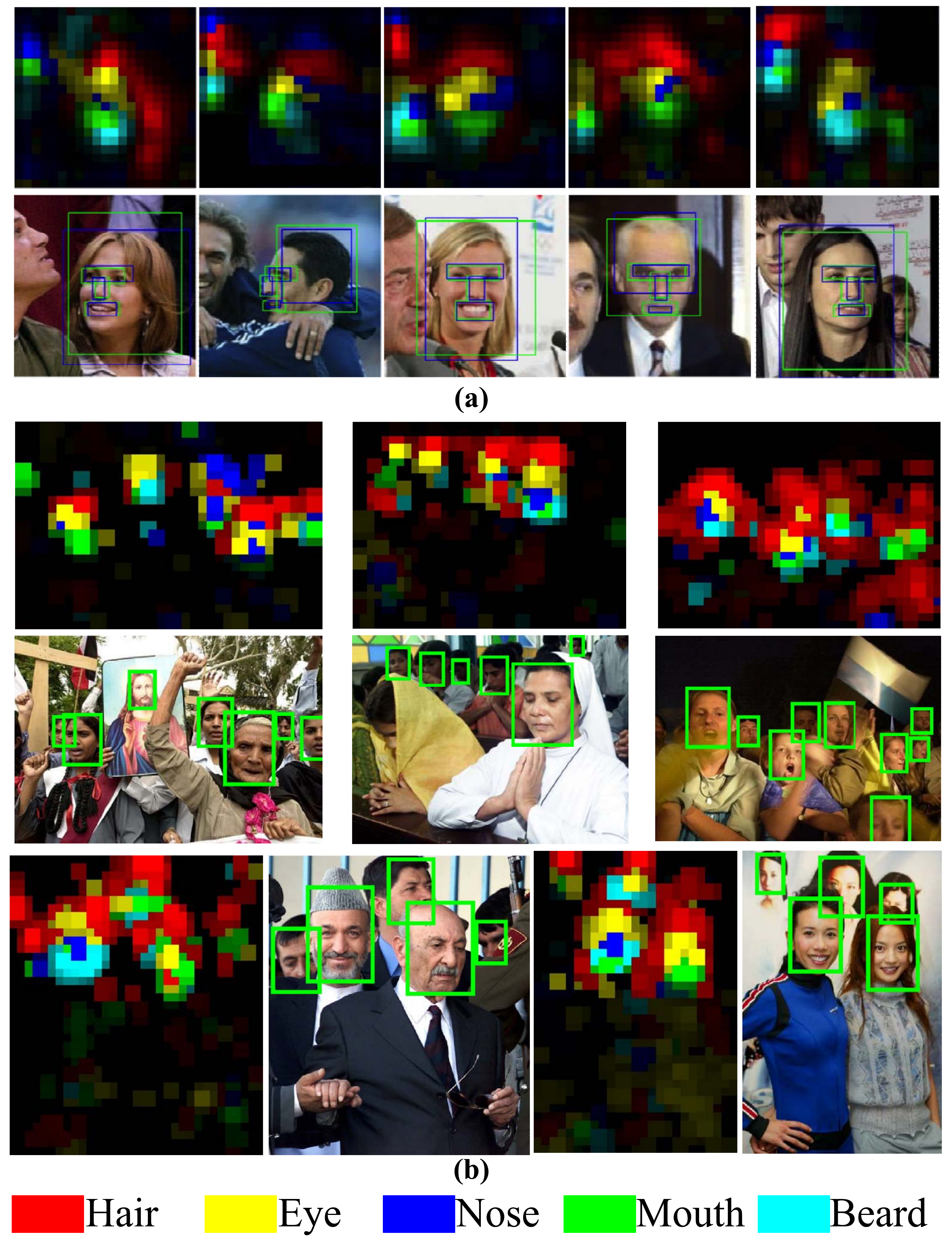}
 {\caption{\small{(a) The top row depicts the response maps generated by the proposed approach on each part. The second row shows the part localization results. Ground truth is depicted by the blue bounding boxes, whilst our part proposals are indicated in green.
(b) Face detection results on FDDB images. The bounding box in green is detected by our method. We show the partness maps as reference.
}}
\label{fig:vs_ex2}}
\vspace{-0.45cm}
\end{center}
\end{figure}

To support this statement, we conduct an experiment by fine-tuning two different CNNs as in Fig.~\ref{fig:pipeline_1}(a), each of which taking different inputs: (1) uncropped images, which may include large portion of background clutters apart the face; and (2) cropped images, which encompass roughly the face and shoulder regions.
The performance is measured based on the part detection rate\footnote{The face part bounding box is generated by first conducting non-maximum suppression (NMS) on the partness maps, and finding bounding boxes centered on NMS points.}.
Note that we combine the evaluation on `Hair+Beard' to suit the ground truth provided by~\cite{crf} (see Sec.~\ref{sec:settings}).
The detection results are summarized in Table~\ref{tab:acc_face_localization}.
As can be observed, the proposed approach performs similarly given both the uncropped and cropped images as training inputs.
The results suggest the robustness of the method in handling unconstrained images for training. In particular, thanks to the facial attribute-driven training, despite the use of uncropped images, the deep model is encouraged to discover and capture the facial part representation in the deep layers, it is therefore capable of generating response maps that precisely pinpoint the locations of parts.
In the following experiments, all the proposed models are trained on uncropped images. Fig.~\ref{fig:vs_ex2}(a) shows the qualitative results. Note that facial parts can be discovered despite challenging poses.

\begin{table}[t]
\begin{center}
\caption{Facial part detection rate. The number of proposals = 350.}
\label{tab:acc_face_localization}
\vskip 0.15cm
\small
\addtolength{\tabcolsep}{-1pt}
\begin{tabular}{c|c|c|c|c}
\hline
\textbf{Training Data} & \textbf{Hair}& \textbf{Eye}& \textbf{Nose}& \textbf{Mouth}
\\
\hline\hline
Cropped& $95.56\%$& $95.87\%$& $92.09\%$& $94.17\%$\\
Uncropped& $94.57\%$& $97.19\%$& $91.25\%$& $93.55\%$\\
\hline
\end{tabular}
\end{center}
\vspace{-0.6cm}
\end{table}

\subsection{From Part Responses to Face Proposal}
\label{sec:evaluate_face_proposal}

\noindent \textbf{Comparison with generic object proposals.}
In this experiment, we show the effectiveness of adapting different generic object proposal generators~\cite{arbelaez2014multiscale, zitnick2014edge, uijlings2013selective} to produce face-specific proposals. Since the notion of face proposal is new, no suitable methods are comparable therefore we use the original generic methods as baselines.
We first apply any object proposal generator to generate the proposals and we use our method described in Sec.~\ref{sec:faceness_measure} to obtain the face proposals.
We experiment with different parameters for the generic methods, and choose parameters that produce moderate number of proposals with very high recall.
Evaluation is conducted following the standard protocol~\cite{zitnick2014edge}.

\begin{figure}[t]
\begin{center}
\includegraphics[width=\linewidth]{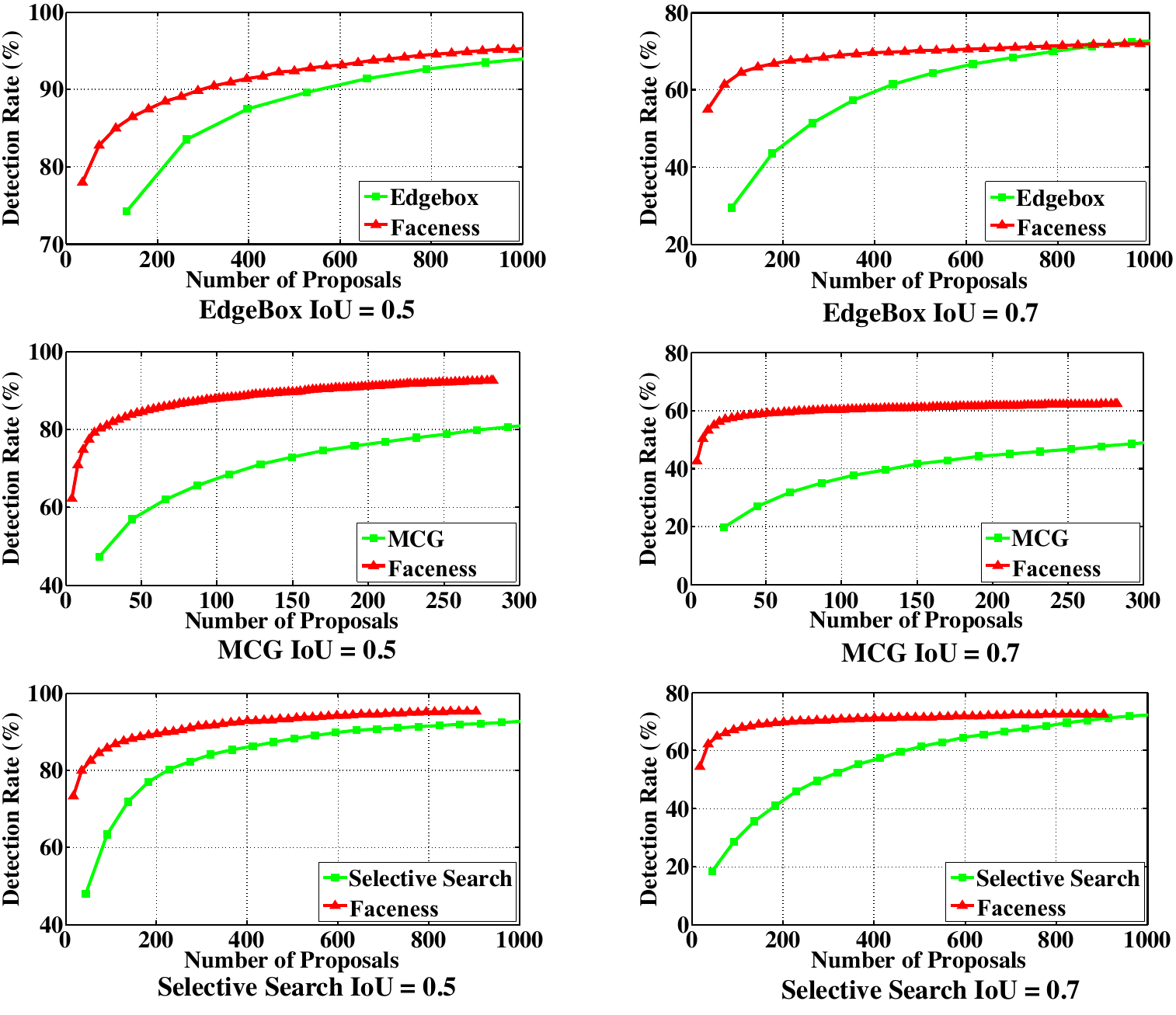}
\vskip -0.2cm
\caption{\small{Comparing the performance between the proposed faceness measure and various generic objectness measures on proposing face candidate windows.}}
\vskip -0.25cm
\label{fig:compare_generic_proposal}
\end{center}
\end{figure}

The results are shown in Fig.~\ref{fig:compare_generic_proposal}. It can be observed that our method consistently improves the state-of-the-art methods for proposing face candidate windows, under different IoU thresholds.
Table~\ref{tab:compare_generic_proposal} shows that our method achieves high recall with small number of proposals.

\begin{figure}[t]
\begin{center}
\subfigure[]{
	 \includegraphics[width=0.52\linewidth]{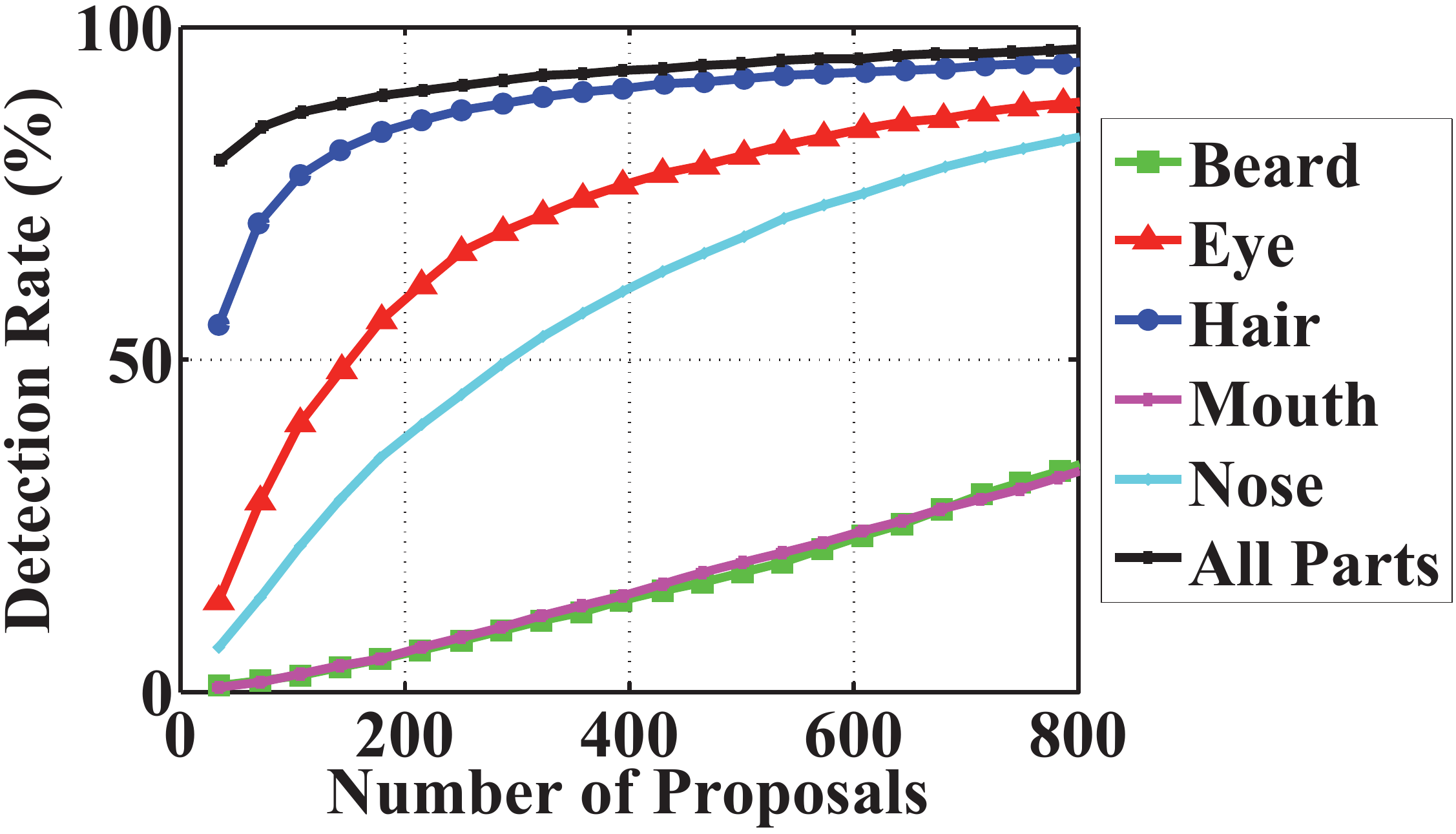}
}
\subfigure[]{
	 \includegraphics[width=0.42\linewidth]{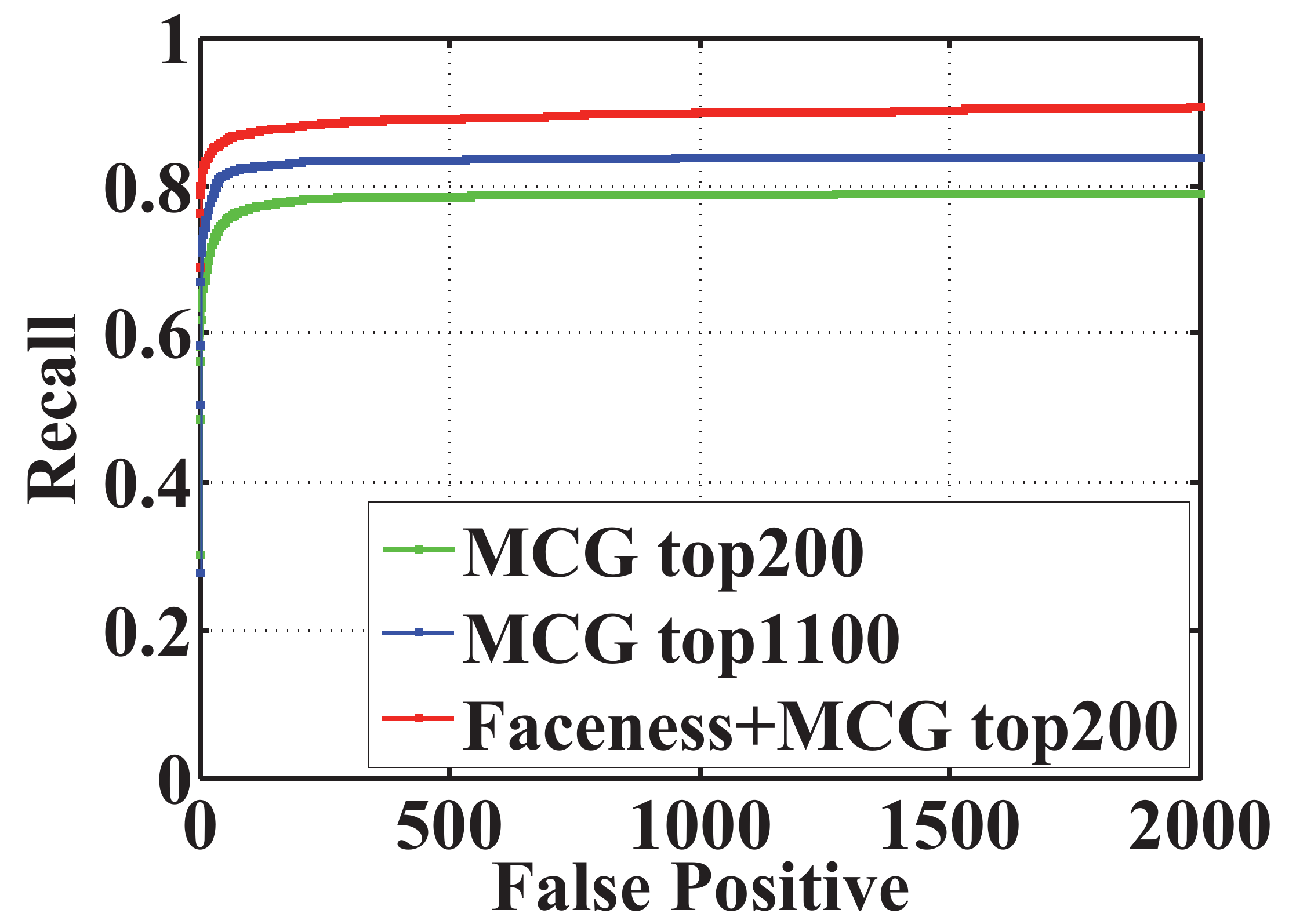}
}
\caption{\small{(a) Contribution of different face parts on face proposal. (b) FDDB face detection results with different proposal methods.}}
\vskip -0.45cm
\label{fig:evaluate_part_contribution}
\end{center}
\end{figure}

\noindent \textbf{Evaluate the contribution of each face part.}
We factor the contributions of different face parts to face proposal.
Specifically, we generate face proposals with partness maps from each face part individually using the same evaluation protocol in previous experiment.
As can be observed from Fig.~\ref{fig:evaluate_part_contribution}(a), the hair, eye, and nose parts perform much better than mouth and beard.
The lower part of the face is often occluded, making the mouth and beard less effective in proposing face windows.
In contrast, hair, eye, and nose are visible in most cases. Nonetheless, mouth and beard can provide complementary cues.

\begin{table}[t]
\begin{center}
\caption{The number of proposals needed for different recalls.}
\label{tab:compare_generic_proposal}
\vspace{0.1cm}
\scriptsize
\addtolength{\tabcolsep}{-1pt}
\begin{tabular}{c|c|c|c|c}
\hline
\textbf{Proposal method} & \textbf{$75\%$}& \textbf{$80\%$}& \textbf{$85\%$}& \textbf{$90\%$}
\\
\hline\hline
EdgeBox~\cite{zitnick2014edge}& $132$& $214$& $326$& $600$\\
EdgeBox~\cite{zitnick2014edge}+Faceness& $\mathbf{21}$& $\mathbf{47}$& $\mathbf{99}$& $\mathbf{288}$ \\
\hline
MCG~\cite{arbelaez2014multiscale}& $191$& $292$& $453$& $942$\\
MCG~\cite{arbelaez2014multiscale}+Faceness& $\mathbf{13}$& $\mathbf{23}$& $\mathbf{55}$& $\mathbf{158}$ \\
\hline
Selective Search~\cite{uijlings2013selective}& $153$& $228$& $366$& $641$ \\
Selective Search~\cite{uijlings2013selective}+Faceness& $\mathbf{24}$& $\mathbf{41}$& $\mathbf{91}$& $\mathbf{237}$ \\
\hline
\end{tabular}
\end{center}
\vspace{-0.5cm}
\end{table}

\begin{figure}[t]
\begin{center}
\includegraphics[width=0.7\linewidth]{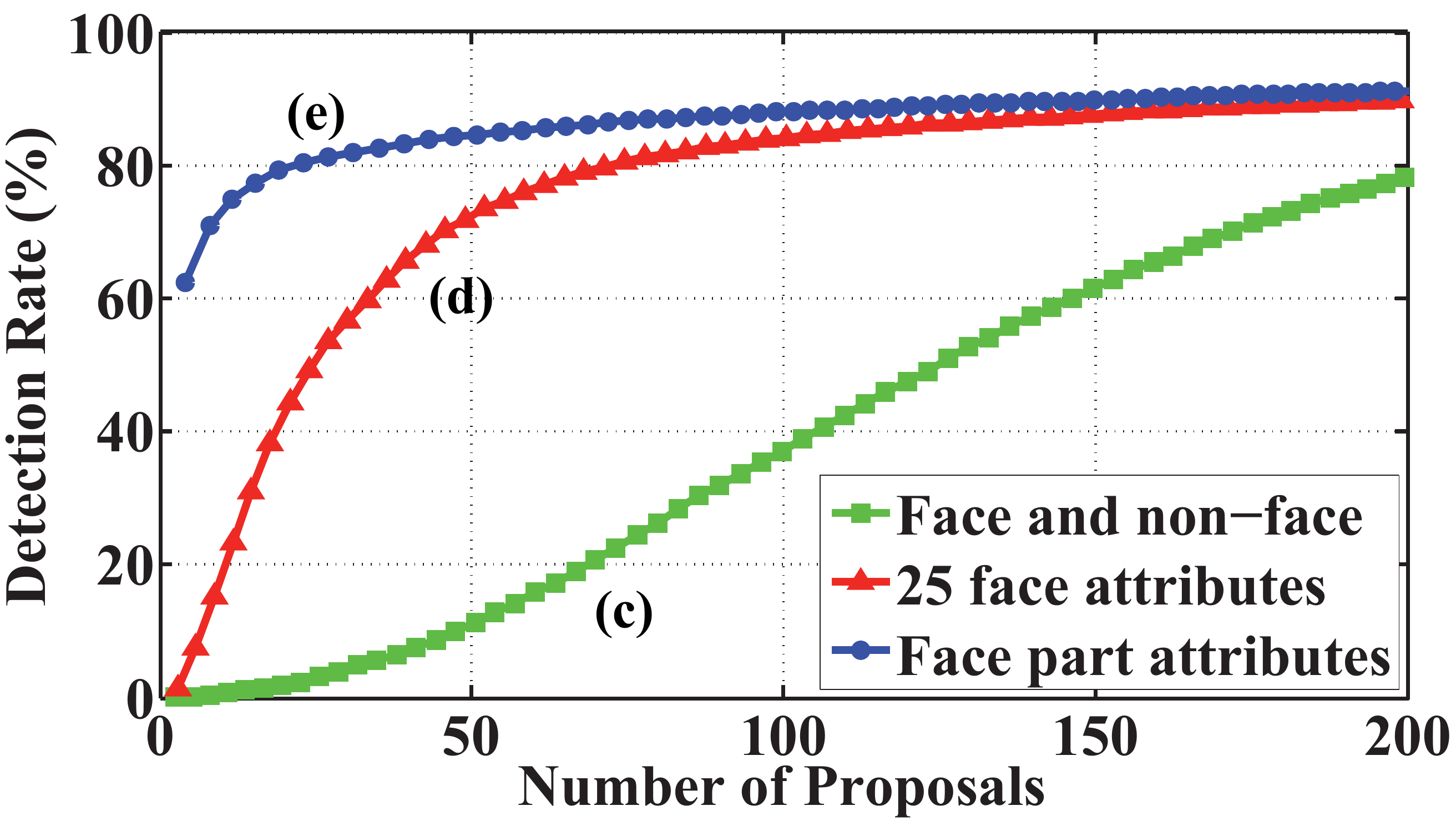}
\caption{Comparing face proposal performance between different training strategies. Methods (c)-(e) are similar to those in Fig.~\ref{fig:partness_map_learning}. Method (e) is our approach.}
\label{fig:quantitative_training_strategy}
\vspace{-0.45cm}
\end{center}
\end{figure}

\noindent
\textbf{Face proposals with different training strategies.}
As discussed in Sec.~\ref{sec:partness}, there are different fine-tuning strategies that can be considered for generating a response map. We compare face proposal performance between different training strategies. 
Quantitative results in Fig.~\ref{fig:quantitative_training_strategy} shows that our approach performs significantly better than approaches (c) and (d). This suggests that attributes-driven fine-tuning is more effective than `face and non-face' supervision.
As can be observed in Fig.~\ref{fig:partness_map_learning} our method generates strong response even on the occluded face compared with approach (d), which leads to higher quality of face proposal.

\begin{figure}[t]
\begin{center}
\includegraphics[width=1\linewidth]{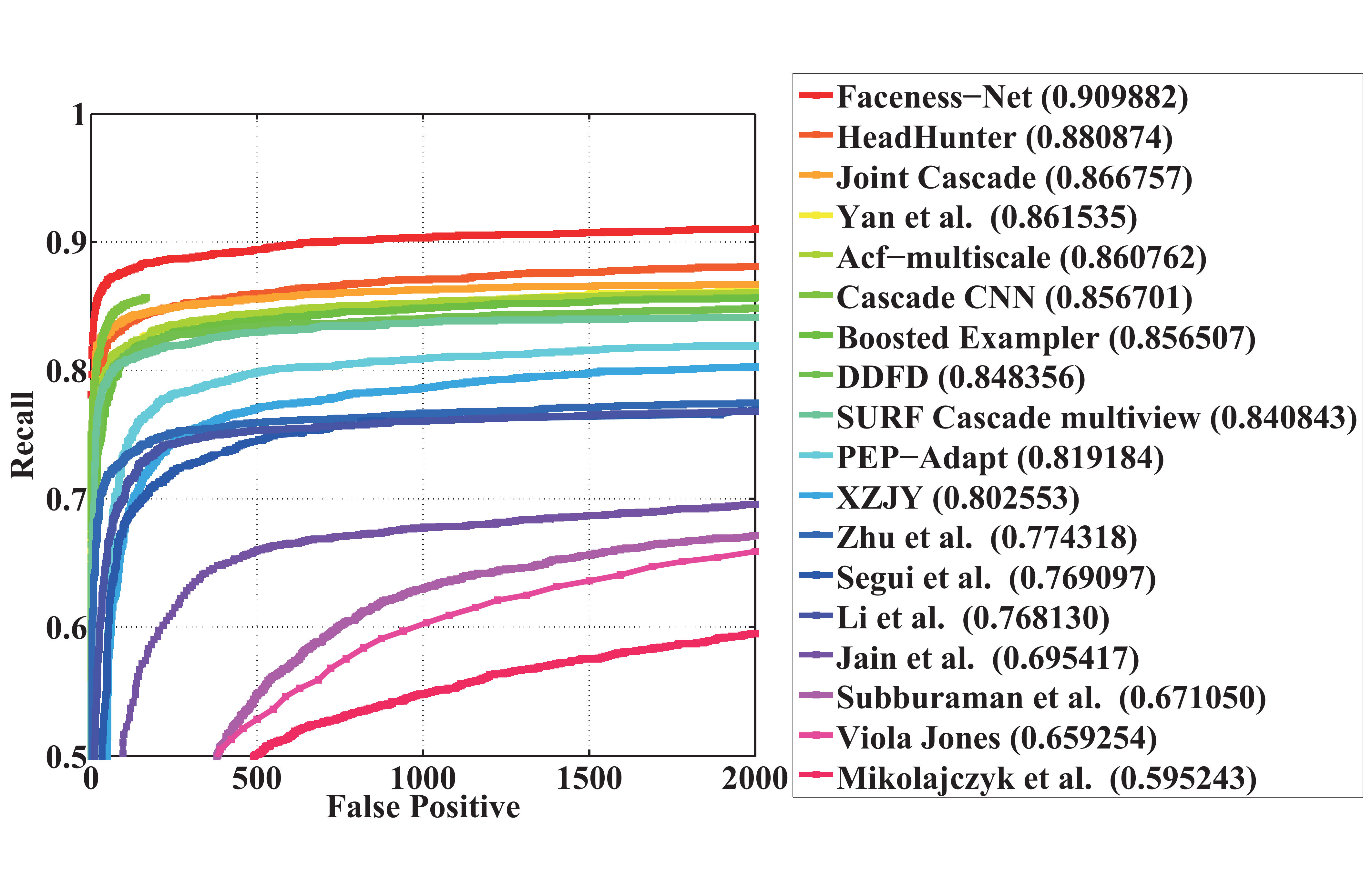}
\vskip -0.1cm
\caption{\small{FDDB results. Recall rate is shown in the parenthesis.}}
\vskip -0.5cm
\label{fig:fddb_results}
\end{center}
\vspace{-0.3cm}
\end{figure}

\begin{figure}[t]
\begin{center}
\includegraphics[width=1\linewidth]{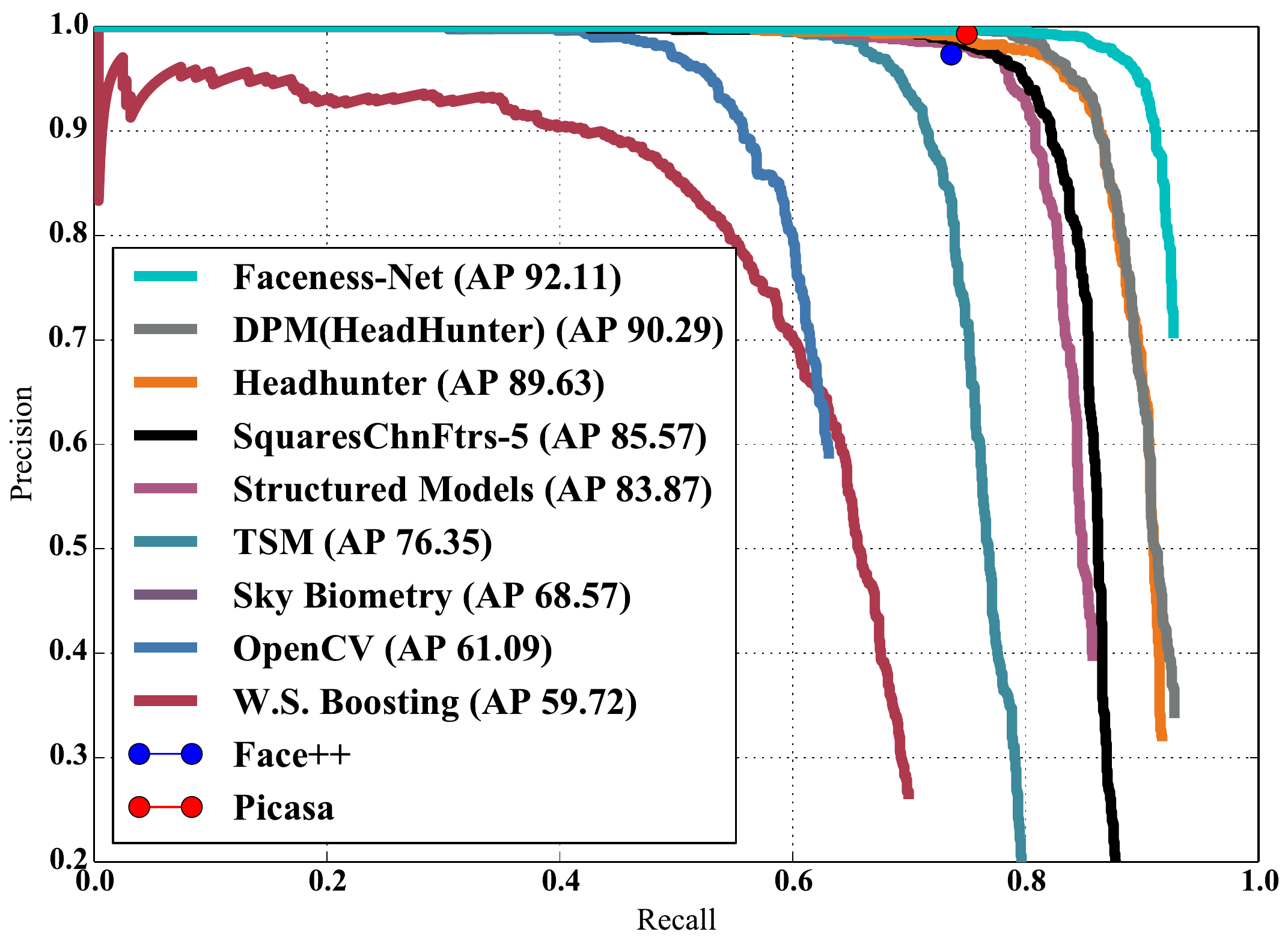}
\vskip -0.25cm
\caption{\small{Precision-recall curves on PASCAL faces dataset. AP = average precision.}}
\vskip -0.5cm
\label{fig:pascal_results}
\end{center}
\vspace{-0.3cm}
\end{figure}

\begin{figure}[t]
\begin{center}
\includegraphics[width=1\linewidth]{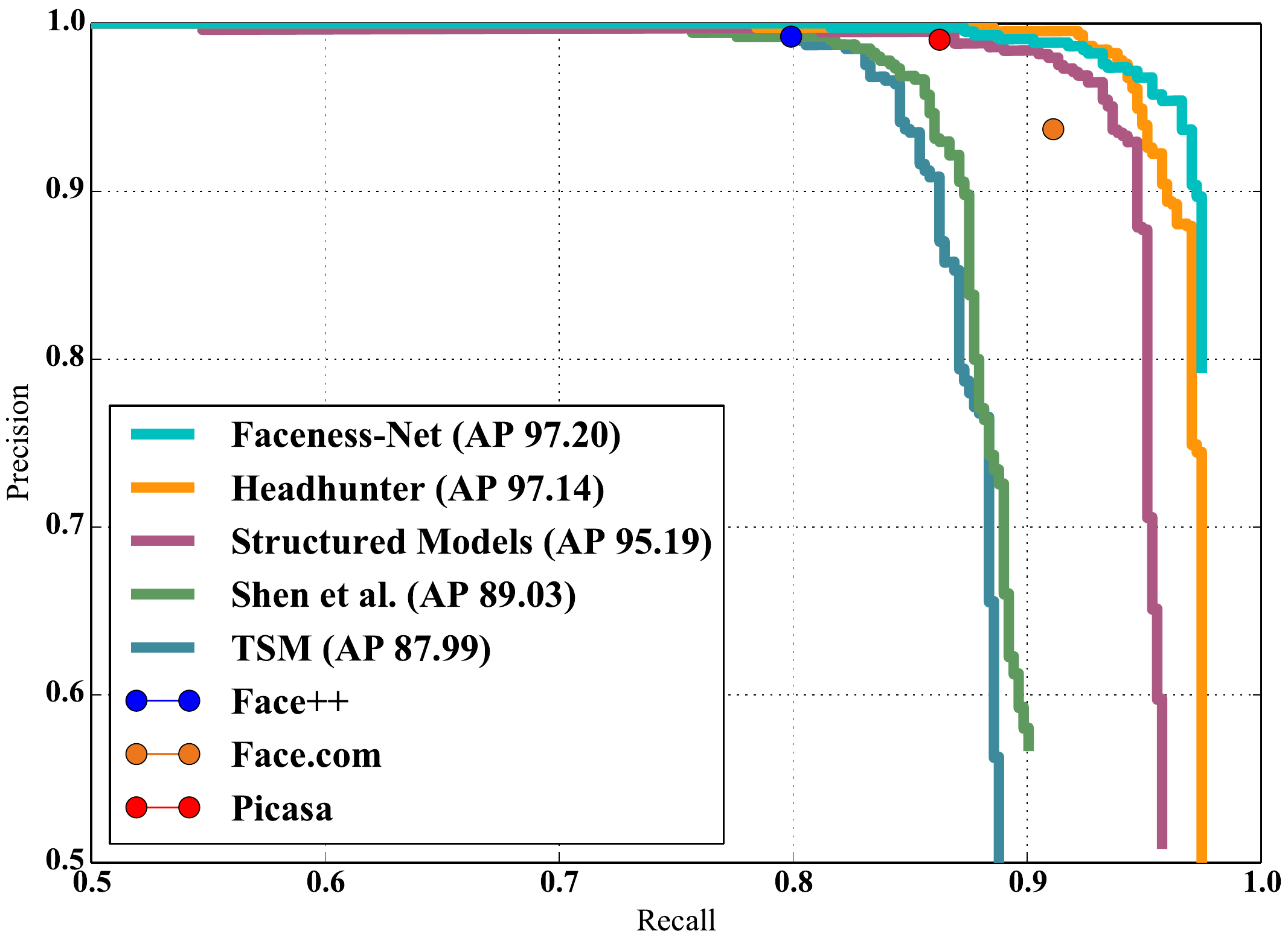}
\vskip -0.25cm
\caption{\small{Precision-recall curves on AFW dataset. AP = average precision.}}
\vskip -0.5cm
\label{fig:afw_results}
\end{center}
\vspace{-0.3cm}
\end{figure}

\begin{figure*}[t]
\begin{center}
{\includegraphics[width=1\linewidth]{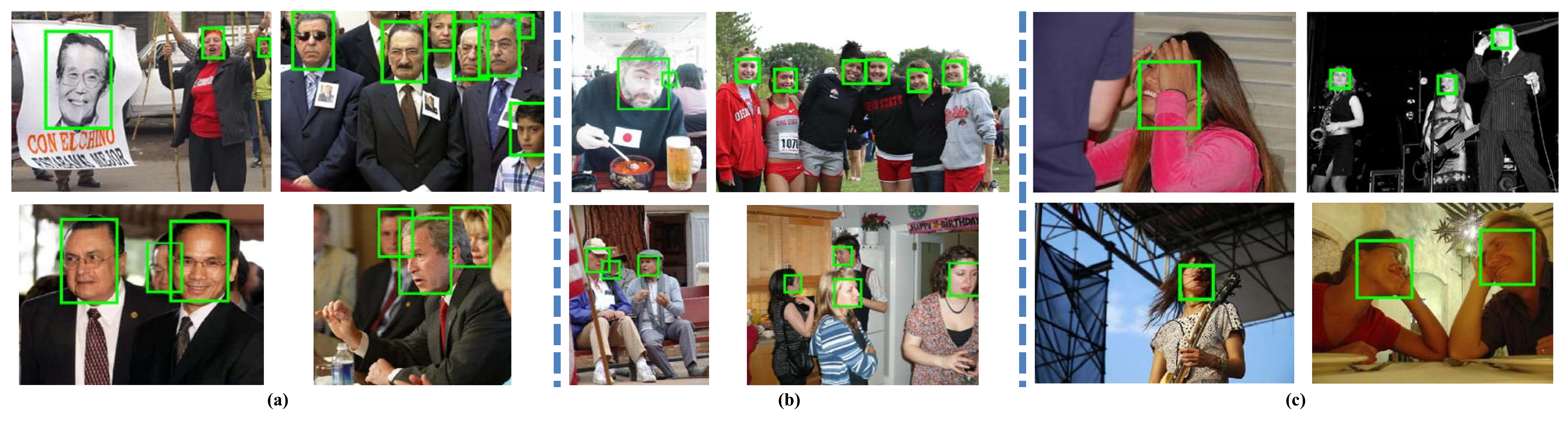}}
\vskip -0.2cm
{\caption{\small{Qualitative face detection results by Faceness-Net on FDDB (a), AFW (b), PASCAL faces (c).}}\label{fig:qualitative_result}}
\vspace{-0.45cm}
\end{center}
\end{figure*}

\subsection{From Face Proposal to Face Detection}
\label{sec:evaluate_face_detection}
In this experiment, we first show the influence of training a face detector using generic object proposals and our face proposals. Next we compare our face detector, Faceness-Net, with state-of-the-art face detection approaches.

\noindent \textbf{Generic object proposal versus face proposal.} We choose the best performer in Fig.~\ref{fig:compare_generic_proposal}, \ie~MCG, to conduct this comparison.
The result is shown in Fig.~\ref{fig:evaluate_part_contribution}(b). The best performance, a recall of $93\%$, is achieved by using our faceness measure to re-rank the MCG top $200$ proposals (Faceness+MCG top-$200$).
Using MCG top $200$ proposals alone yields the worst result. Even if we adjust the number of MCG proposal to $1,100$ with a high recall rate similar to that of our method, the result is still inferior due to the enormous number of false positives. The results suggest that the face proposal generated by our approach is more accurate in finding faces than generic object proposals for face detection.

\noindent \textbf{Comparison with face detectors.}
We conduct face detection experiment on three datasets FDDB~\cite{fddb}, AFW~\cite{zhu2012face} and PASCAL faces~\cite{yan2014face}. Our face detector, Faceness-Net, is trained with top $200$ proposals by re-ranking MCG proposals following the process described in Sec.~\ref{sec:face_detection}. We adopt the PASCAL VOC precision-recall protocol for evaluation.

We compare Faceness-Net against all published methods~\cite{ACF-multiscale,HeadHunter,JointCascade,FastDPM,BoostedExemplar,SURF,PEP-Adapt,XZJY,zhu2012face,VJain} in the FDDB.
For the PASCAL faces and AFW we compare with (1) deformable part based methods, \eg~ structure model~\cite{yan2014face} and Tree Parts Model (TSM)~\cite{zhu2012face}; (2) cascade-based methods, \eg~Headhunter~\cite{HeadHunter}.
Figures~\ref{fig:fddb_results},~\ref{fig:pascal_results}, and~\ref{fig:afw_results} show that Faceness-Net outperforms all previous approaches by a considerable margin, especially on the FDDB dataset. Fig~\ref{fig:vs_ex2}(b) shows some qualitative results on FDDB dataset together with the partness maps. More detection results are shown in Fig~\ref{fig:qualitative_result}.

\section{Discussion}

There is a recent and concurrent study that proposed a Cascade-CNN~\cite{cascadecnn} for face detection. Our method differs significantly to this method in that we explicitly handle partial occlusion by inferring face likeliness through part responses. This difference leads to a significant margin of $2.65\%$ in recall rate (Cascade-CNN $85.67\%$, our method $88.32\%$) when the number of false positives is fixed at 167 on the FDDB dataset. The complete recall rate of the proposed Faceness-Net is $90.99\%$ compared to $85.67\%$ of Cascade-CNN.

At the expense of recall rate, the fast version of Cascade-CNN achieves $14$fps on CPU and $100$fps on GPU for $640\times480$ VGA images.
The fast version of the proposed Faceness-Net can also achieve practical runtime efficiency, but still with a higher recall rate than the Cascade-CNN.
The speed up of our method is achieved in two ways. First, we share the layers from conv1 to conv5 in the first stage of our model since the face part responses are only captured in layer conv7 (Fig.~\ref{fig:pipeline_1}). The computations below conv7 in the ensemble are mostly redundant, since their filters capture global information \eg~edges and regions.
Second, to achieve further efficiency, we replace MCG with Edgebox for faster generic object proposal, and reduce the number of proposal to $150$ per image. Under this aggressive setting, our method still achieves a $87\%$ recall rate on FDDB, higher than the $85.67\%$ achieved by the full Cascade-CNN.
The new runtime of our two-stage model is $50$ms on a single GPU\footnote{We use the same Nvidia Titan Black GPU as in Cascade-CNN~\cite{cascadecnn}.} for VGA images. The runtime speed of our method is comparatively lower than~\cite{cascadecnn} because our implementation is currently based on unoptimized MATLAB code.

We note that further speed-up is possible without much trade-off on detection performance. Specifically, our method will benefit from Jaderberg~\etal~\cite{Speeding_up_CNN}, who show that a CNN structure can enjoy a $2.5\times$ speedup with no loss in accuracy by approximating non-linear filtering with low-rank expansions. Our method will also benefit from the recent model compression technique~\cite{DarkKnowledge}.

\small{\textbf{Acknowledgement} This work\footnote{For more technical details, please contact the corresponding author Ping Luo via pluo.lhi@gmail.com.} is partially supported by the National Natural Science Foundation of China (91320101, 61472410, 61503366).}

{\small
\bibliographystyle{ieee}
\bibliography{faceness_final,short}
}

\end{document}